\documentclass[preprint,12pt]{elsarticle}

\usepackage{hyperref}
\usepackage{amsmath}
\usepackage[capitalise]{cleveref}
\usepackage{graphicx}
\usepackage{graphics}
\usepackage{subfig}
\usepackage{color}
\usepackage{multicol}
\usepackage{svg}
\usepackage{todonotes}
\usepackage{bbm}
\usepackage{cleveref}
\usepackage{setspace}

\usepackage[left = 1in, right = 1in, top = 1in]{geometry}
\usepackage{amssymb}

\usepackage{lineno}

\biboptions{authoryear}
\makeatletter
\def\ps@pprintTitle{%
  \let\@oddhead\@empty
  \let\@evenhead\@empty
  \let\@oddfoot\@empty
  \let\@evenfoot\@oddfoot
}
\makeatother
\begin{document}

\begin{frontmatter}
\journal{Sustainable Cities and Society}
\title{Estimating city-wide hourly bicycle flow using a hybrid LSTM MDN}

\author[DTU]{Marcus Skyum Myhrmann
\corref{mycorrespondingauthor}}
\cortext[mycorrespondingauthor]{Corresponding author. Tel: +4542956003, Current adress: Technical University of Denmark, Bygningstorvet Building 116B, Kgs. Lyngby, 2800, Denmark}
\ead{mskyum@dtu.dk}

\author[DTU]{Stefan Eriksen Mabit}
\ead{smab@dtu.dk}
\address[DTU]{Transport Division, DTU Management, Technical University of Denmark, Bygningstorvet Building 116B, Kgs. Lyngby, 2800, Denmark}

\begin{abstract}
Cycling can reduce greenhouse gas emissions and air pollution and increase public health. With this in mind, policy-makers in cities worldwide seek to improve the bicycle mode-share. However, they often struggle against the fear and the perceived riskiness of cycling. Efforts to increase the bicycle's mode-share involve many measures, one of them being the improvement of cycling safety. This requires the analysis of the factors surrounding accidents and the outcome. However, meaningful analysis of cycling safety requires accurate bicycle flow data that is generally sparse or not even available at a segment level. 
Therefore, safety engineers often rely on aggregated variables or calibration factors that fail to account for variations in the cycling traffic caused by external factors. 

This paper fills this gap by presenting a Deep Learning based approach, the Long Short-Term Memory Mixture Density Network (LSTMMDN), to estimate hourly bicycle flow in Copenhagen, conditional on weather, temporal and road conditions at the segment level. This method addresses the shortcomings in the calibration factor method and results in $66-77\%$ more accurate bicycle traffic estimates.

To quantify the impact of more accurate bicycle traffic estimates in cycling safety analysis,  we estimate bicycle crash risk models to evaluate bicycle crashes in Copenhagen. The models are identical except for the exposure variables being used. One model is estimated using the LSTMMDN estimates, one using the calibration-based estimates, and one using yearly mean traffic estimates. The results show that investing in more advanced methods for obtaining bicycle volume estimates can benefit the quality, mitigating efforts by improving safety analyses and other performance measures.
\end{abstract}


\end{frontmatter}  

\section{Introduction}
In a world where sustainable transport is increasingly essential, policy-makers in cities seek to increase the mode share of the bicycle. Not only is cycling emission-free, but it also improves the health and wellbeing of its users \citep{Mueller2015HealthReview} and leads to the improved livability of cities. 
However, a frequently reported barrier to increasing the mode share of cyclists is the fear of traffic-related injury \citep{Horton2016FearCycling, TransportforLondon2014AttitudesTfL, Vejdirektoratet2018HvorforCyklen}.

To tackle the above problems, safety engineers, transport agencies, and researchers have investigated various aspects of bicycle accidents to identify the factors associated with bicycle crash occurrence \citep{Boele-Vos2017CrashesStudy, Janstrup2019APrioritization, Aldred2018CyclingLimits, Vandenbulcke2014PredictingApproach, Dozza2017CrashData, Rossetti2018ModelingInfrastructure, Twisk2013AnUse, Morrison2019On-roadCrashes, Kaplan2015ARegion,Ji2021GeographicallyStudy, Saha2018SpatialModels, Raihan2019EstimationModels}, and injury outcome\citep{Myhrmann2021FactorsCrashes, Kaplan2014AggravatingDenmark, Fountas2021AddressingMeans, Kim2007BicyclistAccidents, Behnood2014LatentSeverities, Thomas2013TheLiterature, Chen2017HowAnalysis, Samerei2021UsingCrashes} to make informed mitigating efforts.
However, as highlighted by \citet{Dozza2017CrashData} and \citet{Thomas2013TheLiterature}, many such investigations do not account for cyclist exposure. This is primarily due to a lack of exposure data, as bicycle monitoring is often infrequent, only conducted at a few locations, or even entirely unavailable. 
The studies that account for cyclist exposure often rely on highly aggregated exposure measures such as the annual average daily cycling traffic (AADCT), annual average weekday cycling traffic (AAWCT), or population and commuter indicators.
Transport agencies use calibration factors applied to AADCT or AAWCT to obtain reasonable hourly volume profiles, \citep{Schrank20212021Methodology}. However, the hourly cycling volumes derived from calibration factors do not reflect variations in traffic due to weather, temporal effects, and other external factors. This presents a significant issue considering the impact weather and other factors have on cyclist ridership \citep{Bocker2013ImpactReview, Nankervis1999TheCommuting, Nosal2014TheCounts}.

Since detailed traffic volume data are essential to provide the most relevant accident analyses \citep{Norros2016}, there is a need for improved bicycle volume estimates based on the limited data sources available.

Recent advances for the estimation of traffic volumes have primarily focused on the prediction of the short-term traffic state/flow \citep{Lv2015TrafficApproach, Polson2017DeepPrediction, Du2021AnAlgorithm, Chen2018ForecastingAlgorithm}. Short-term traffic prediction is mainly relevant for segments and networks already subject to good monitoring, where detailed short-term forecasts can be applicable for congestion easing. This is because the short-term traffic predictions are conditioned on the previous traffic flow data. An especially popular model to aid this task is the Long Short-Term Memory (LSTM) neural network \citep{Hochreiter1997LongMemory} which has been adopted for many recent models for traffic forecasting \citep{Ma2015LongData, Duan2016TravelNetwork, Chen2016LongData, Cui2020StackedValues, Zhao2017LSTMForecast}. 

However, to aid safety engineers in improving bicycle accidents and safety analysis, historic bicycle flow estimates are needed. 
Therefore our study focuses on improving the estimation of historical bicycle volumes derived from mean daily exposure measures, the estimation of which are the focus of recently developed large scale models \citep{AledDavies2017Cynemon-CyclingPlanning, Kjems2019COMPASS:Hovedstadsomradet}. The model should estimate historic bicycle flow unconditional of previous traffic flow, where only the mean-expected daily traffic is available.

To accomplish this, we apply a novel neural network approach to estimate hourly bicycle volumes conditional on weather conditions, temporal effects, and road conditions to overcome the calibration factor method's shortcomings. This framework is a hybrid of an LSTM and a Mixture Density Network \citep{Bishop1994MixtureNetworks}, which introduces a Gaussian mixture model (GMM) extension to the traditional LSTM. This hybridisation enables the model to estimate a conditional bicycle flow distribution in contrast to the conventional conditional mean estimation. Furthermore, the MDN extension improves the estimates of hourly bicycle volumes by treating them as random draws from a distribution, thus introducing variation across the network even on measurably similar roads.
Finally, we quantify the effect of improved bicycle exposure estimates by contrasting crash frequency models using different exposure variables.

\section{Methodology} 
The following section describes the methodological approach employed to estimate the bicycle flow. 
\subsection{LSTM}
\begin{figure}[htb]
    \centering
    \includegraphics[width = 10cm, trim=0 3cm 0 0, clip]{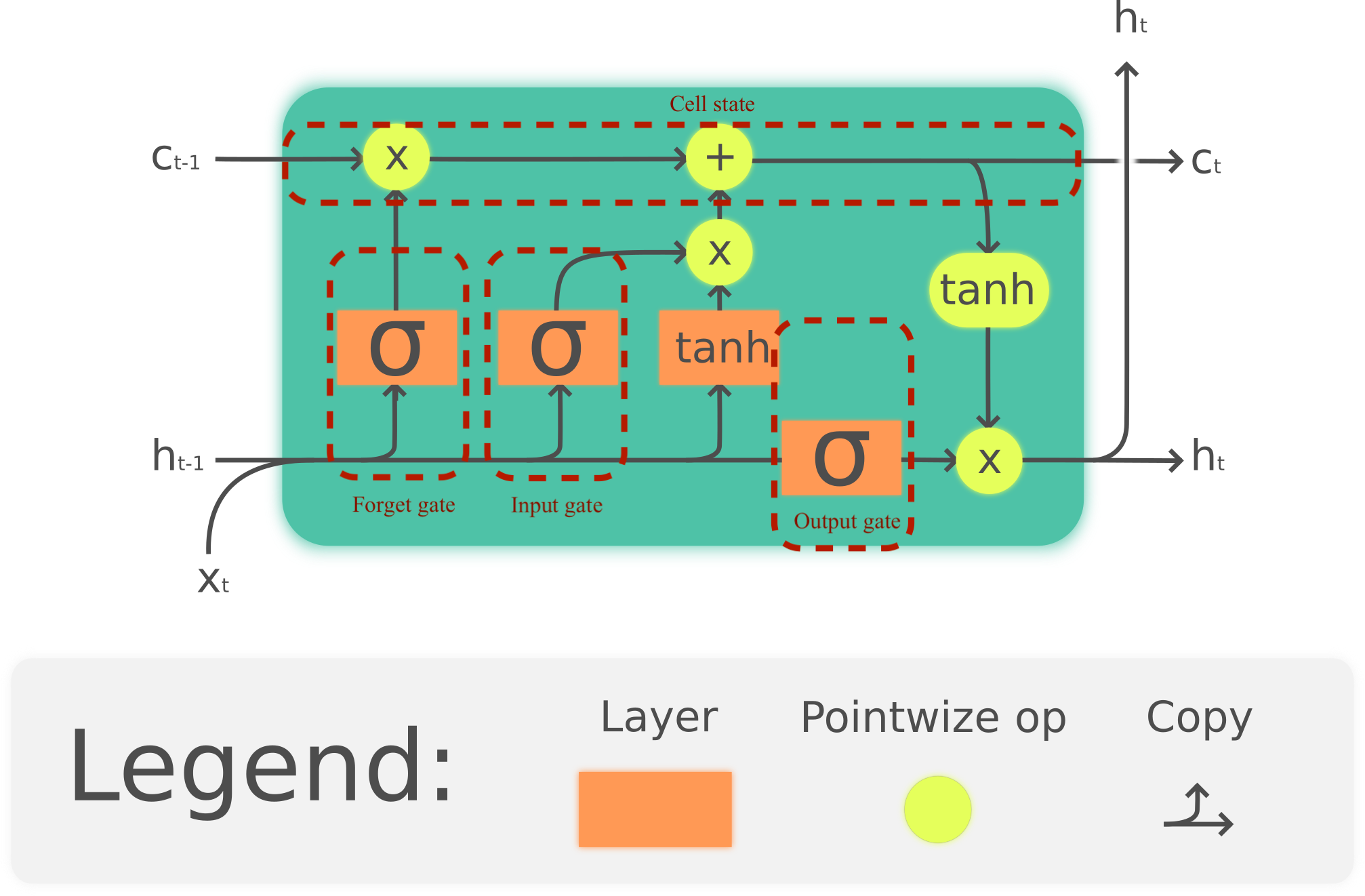}
    \caption[Caption for LOF]{Graphical illustration of LSTM cell structure\footnotemark: Orange squares mark layers with respective activation functions, yellow circles mark point-wise operation, where x and + icons indicate respective element-wise multiplication and addition of tensors in the LSTM cell.}
    \label{fig:LSTM}
\end{figure}

LSTMs have exhibited a superior capability of handling nonlinear time series problems \citep{Hochreiter1997LongMemory}.  
The LSTM learns to represent temporal data by introducing a memory cell and sub-processes, referred to as gates. There are three such gates in the LSTM cell: the input gate, the forget gate, and the output gate. The gates each handle different tasks, involving what information to keep from the previous cell state, what new input to consider, and which to include in the cell state. In \Cref{fig:LSTM} there is a visual illustration of the LSTM cell layout, and the computations performed in the LSTM cell for each time step in the temporal sequence are shown in \Crefrange{eq:forget}{eq:hidden}.

\footnotetext{https://upload.wikimedia.org/wikipedia/commons/3/3b/The\_LSTM\_cell.png}

\begin{align}
    f_{t} &= \sigma(W_{fx}x_{t} + W_{fh}h_{t-1} + b_{f})\label{eq:forget}\\
    i_{t} &= \sigma(W_{ix}x_{t} + W_{ih}h_{t-1} + b_{i})\label{eq:input}\\
    \tilde{C_{t}} &= tanh(W_{Cx}x_{t} + W_{Ch}h_{t-1} + b_{C})\label{eq:ctilde}\\
    C_{t} &= f_{t}\times C_{t-1} + i_{t}\times\tilde{C_{t}}\label{eq:ct}\\
    o_{t} &= \sigma(W_{ox}x_{t} + W_{oh}h_{t-1} + b_{o})\label{eq:output}\\
    h_{t} &= tanh(C_{t}) \times o_{t} \label{eq:hidden}
\end{align}

Here $\sigma(\cdot)$ and $\tanh(\cdot)$ are the respective activation functions, $W_{fx}, W_{fh}, W_{ix}, W_{ih}, W_{ox}, W_{oh}, W_{Cx}$ and $W_{Ch}$ are the weight matrices of the respective gates $f_{t}, i_{t}, o_{t}$ and the memory cell $C_{t}$, in the LSTM cell. $b_{f}, b_{i}, b_{o}$ and $b_{C}$ are intercept/bias terms of the respective gates, and $h_{t}$ represents the hidden state at the time step $t$. 

\subsection{LSTM MDN}
\begin{figure}[htbp]
    \centering
    \includegraphics[width=10cm]{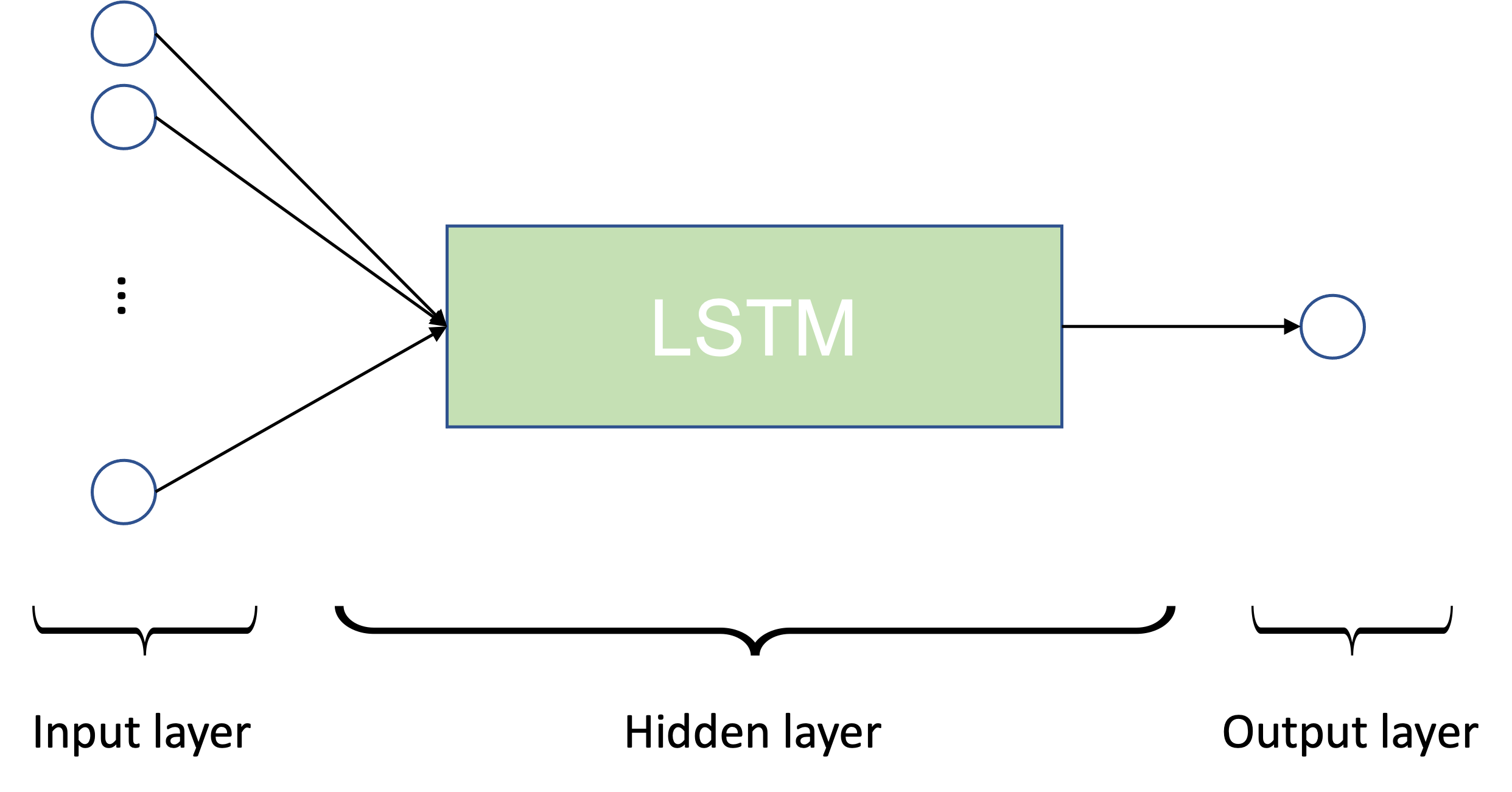}
    \caption{Example structure of simple LSTM neural network}
    \label{fig:LSTMNN}
\end{figure}

An illustration of a simple LSTM regression network is shown in \Cref{fig:LSTMNN}. It has an input layer, a single LSTM cell with $k$ computational nodes in each gate, and a single-node output layer. The optimisation of such a model by minimising the mean squared error (MSE) has been shown to approximate the conditional average of the target data (i.e. bicycle flow) \citep{Bishop1994MixtureNetworks}. 
However, we wish to account for similar weather and seasonal conditions not necessarily yielding the same bicycle flow in this study. Therefore, we introduce a level of randomness in the bicycle flow estimation. Namely, through the Mixture Density Network \citep{Bishop1994MixtureNetworks}.\\

The original Mixture Density Network proposed by \citet{Bishop1994MixtureNetworks} is a combination of an artificial neural network (ANN) and a mixture model \citep{McLachlan1988MixtureClustering}. This combination provides the flexibility to model a general distribution and, as such, enables the estimation of the conditional density function of the target data, in contrast to the conditional average. 

In the mixture model, the probability density of the target data is specified as a linear combination of kernel functions,

\begin{equation}\label{eq:pdf}
p(y|X) = \sum_{i=1}^{A}\alpha_{i}(X) \phi_{i}(y|X)
\end{equation}

where $A$ is the number of mixture components, $\alpha_{i}(X)$ are the mixing coefficients dependent on the input data, and $\phi_{i}(\cdot|X)$ are Gaussian probability density kernels:

\begin{equation}
    \phi_{i}(t|x) \propto \frac{1}{\nu_{i}(x)^{1/2}} e^{-\frac{(t-\mu_{i}(x))^{2}}{2\nu_{i}(x)^{2}}}, \quad i \in C
\end{equation}
where $\mu_{i}(x)$ is the centre of the kernel, i.e. the conditional average, and $\nu_{i}(x)$ the associated variance.

We refer to the combination of the LSTM and the MDN in this study as the LSTMMDN. This model varies from the LSTM regression shown in \Cref{fig:LSTMNN} only in the output layer. The task of the LSTM in the LSTMMDN is to estimate the conditional input to the mixture model, namely the mixing coefficients $\alpha_{i}(X)$, the means $\mu_{i}(X)$ and the variances $\nu_{i}(X)$, conditional on input $X$. 

To ensure the properties of the conditional density of the target data $p(y|X)$, the mixing coefficients $\alpha_{i}(X)$ need to satisfy the constraint in \Cref{eq:constrain}.

\begin{equation} \label{eq:constrain}
    \sum_{i=1}^{A}\alpha_{i}(X)=1
\end{equation}

This constraint can be satisfied by connecting the mixing coefficients to the feed-in network using a softmax/multinomial logit regression \citep{Bishop1994MixtureNetworks}.

\begin{equation}{\label{eq:softmax}}
    \alpha_{i} = \frac{\alpha_{i}(o_{i}^{\alpha})}{\sum_{j}^{A}o_{j}^{\alpha}}
\end{equation}

It is convenient to avoid the conditional variances tending to zero. Therefore we parameterise the conditional variance in terms of the exponential of the network output, \Cref{eq:exp}. This parametrisation also corresponds to choosing an un-informative Bayesian prior in a Bayesian framework, assuming that the output $o_{i}^{\nu}$ has a uniform probability distribution \citep{Bishop1994MixtureNetworks}.

\begin{equation}\label{eq:exp}
    \nu_{i} = e^{o_{i}^{\nu}}
\end{equation}

Finally, the centres/conditional means $\mu_{i}$, shown in \Cref{eq:means}, are represented by location parameters that depend directly on the network outputs.

\begin{equation}\label{eq:means}
    \mu_{i} = o_{i}^{\mu}
\end{equation}

To optimise the weights and biases  $\Theta = \{W_{fx}, W_{fh}, W_{ix}, W_{ih}, W_{ox}, W_{oh}, W_{Cx}, W_{Ch}, b_{f}, b_{i},$ $ b_{o}, b_{C}\}$ we use a method called backpropagation. This is an iterative process in which data are "fed forward" until a prediction is made upon which the deviation/error from the actual result is computed using a loss function. Based on the deviation/error of the result and the individual weights' ($\Theta$) impact on the output, the error is then backpropagated using a gradient scheme to update the individual weights $\Theta$.
For further explanation of the statistics of backpropagation, see, e.g. \citet{Hastie2009NeuralNetworks, Bishop1994MixtureNetworks}. 
The loss function we use for the parameter optimisation is the log-likelihood as shown in \Cref{eq:LL}.

\begin{equation}\label{eq:LL}
   \mathcal{L}\mathcal{L}(\Theta) =  \frac{1}{N}\sum_{j}^{N}-\ln\{ \sum_{i=1}^{A}\alpha_{i}(X_{j}) \phi_{i}(y_{j}|X_{j}) \}
\end{equation}

\begin{figure}[htbp]
    \centering
    \includegraphics[width = 14cm]{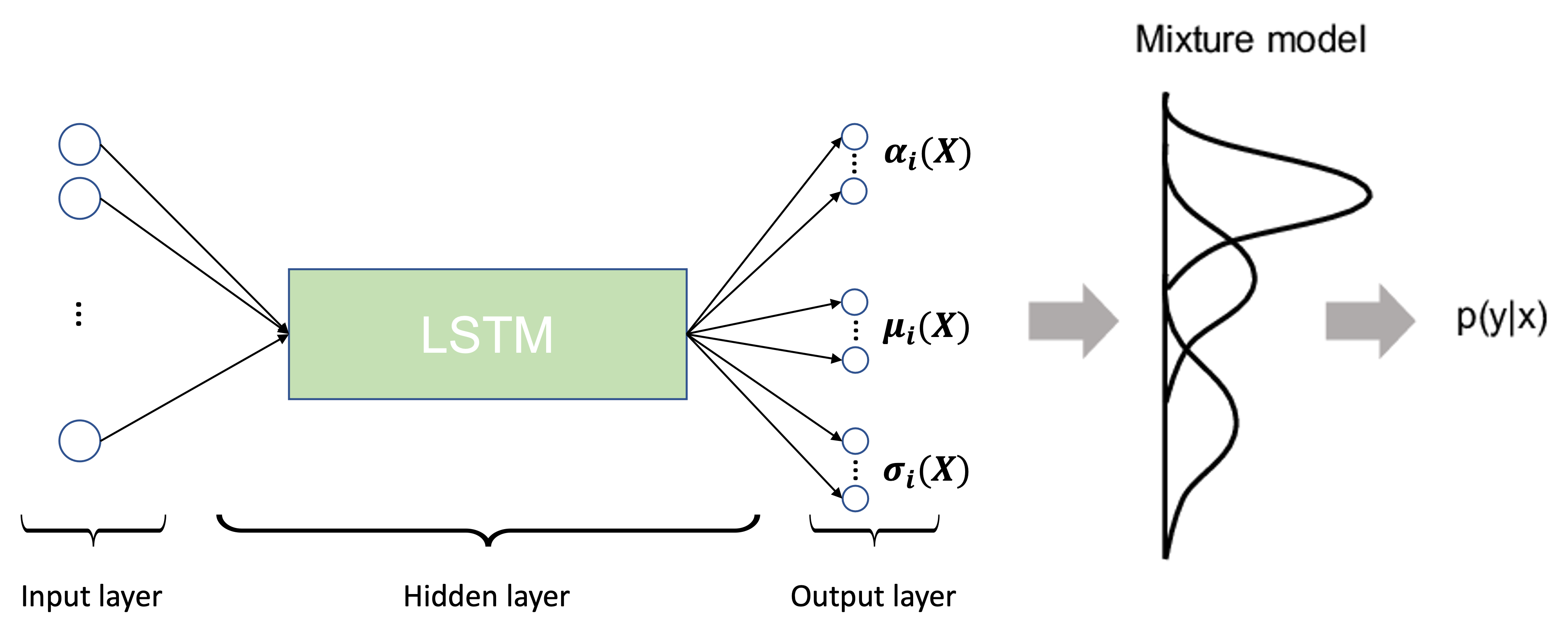}
    \caption{Graphical illustration of the LSTMMDN setup used in the study.}
    \label{fig:LSTMMDN}
\end{figure}

\subsection{Model configuration}
To sum up the model configuration: The LSTMMDN applied in this study is configured as shown in \Cref{fig:LSTMMDN} using the activation functions described in \Crefrange{eq:forget}{eq:hidden}. We apply an internal dropout rate of $20\%$ dropout in the LSTM cell to avoid overfitting/over-specification of the model \citep{Srivastava2014Dropout:Overfitting}. Meanwhile, we use a linear $a(x)=x$ in the output layer, along with the relevant transformations leading into the GMM, as described in \Crefrange{eq:softmax}{eq:means}.
The specific backpropagation scheme used for the parameter optimisation is Adaptive Moment Estimation (Adam) algorithm \citep{Kingma2015Adam:Optimization}. This is a stochastic gradient-based optimisation approach, which provides some advantages over traditional gradient-based schemes \citep{Hastie2009NeuralNetworks}. Finally, we employ an early stopping criterion. This criterion is evaluated on a small validation sample of the data and ends the model estimation if no performance improvement is achieved over $L$ model updates to avoid overfitting. We set $L=100$.

\section{Results}
\subsection{Data and experimental setup}
The current study intends to estimate bicycle flow in Copenhagen, Denmark. 
The hourly bicycle volume data recorded by bicycle counting stations have been acquired for 2017-2020. The bicycle counting stations are placed around Copenhagen and marked by the blue dots in \Cref{fig:meas_st}. There is a considerable variation in active counting days, ranging from 9-731 days, with the median number of active counting days being 40. Overall, a total of 64,664 hourly bicycle volumes are recorded.

The Danish Road Directorate provides the AADCT and AAWCT for each bicycle counting station every year where activity is registered. The AADCT and AAWCT are computed by the Danish Road Directorate and reported along with the bicycle counting data. 

Using the Open Data API of the Danish Meteorological institute \citep{DMI2021DanishConfluence}, located at the green dot in \Cref{fig:meas_st},  we acquired weather data for Copenhagen in the period 2017-2020. The data are reported at 10-minute intervals and contain information on air temperature, pressure, wind speed, wind gusts, wind direction, precipitation levels, visibility and snow volume. These meteorological data are assumed to play a role in cycling ridership and, therefore, are included in the bicycle flow estimation.

Finally, we also include time-related data such as an hour of the day, day of the week, week of the year, and indicators of public holidays, as this would be assumed to influence cyclist ridership strongly. 

The data consist of 64,664 observations of hourly bicycle volumes and $\approx 388,000$ weather and temporal measurements at 10-minute intervals, totalling 17 features. These 17 weather and time-related features, including the AADCT of the roads of the bicycle counters, are the predictors used to estimate the hourly bicycle volumes $y_{T}$. 
The observed 10-minute interval predictors $\pmb{x}_{t}$ are grouped such that a sequence of six 10-minute interval observations $\pmb{x}_{t}$ are paired with the matching response (i.e. the accumulated bicycle volume of the hour) as shown below.

These data are used to train and evaluate the LSTMMDN. The combined data of all bicycle count stations subsequently split into a training, validation and test data set containing $70\%$, $10\%$ and $20\%$ of the total data, respectively. 

\begin{equation}
y_{T}(\pmb{X}_{T} ): \quad \pmb{X}_{T} = 
\begin{pmatrix}
\pmb{x}_{t1}\\
\pmb{x}_{t2}\\
\pmb{x}_{t3}\\
\pmb{x}_{t4}\\
\pmb{x}_{t5}\\
\pmb{x}_{t6}\\
\end{pmatrix} = 
\begin{pmatrix}
(x_{1,t1}, x_{2,t1}, ...., x_{D,t1})\\
(x_{1,t2}, x_{2,t2}, ...., x_{D,t2})\\
(x_{1,t3}, x_{2,t3}, ...., x_{D,t3})\\
(x_{1,t4}, x_{2,t4}, ...., x_{D,t4})\\
(x_{1,t5}, x_{2,t5}, ...., x_{D,t5})\\
(x_{1,t6}, x_{2,t6}, ...., x_{D,t6})
\end{pmatrix}
\end{equation}\label{eq:seq}

Here $t1, t2, t3, t4, t5, t6$ index the six 10-minute intervals of the input sequence for the aggregated hourly cycling volume, $T$ indicates the full hour, and $D$ is the number of predictors. 

\subsubsection{Data pre-processing}
\begin{figure}[htb]
    \centering
    \includegraphics[height=10cm]{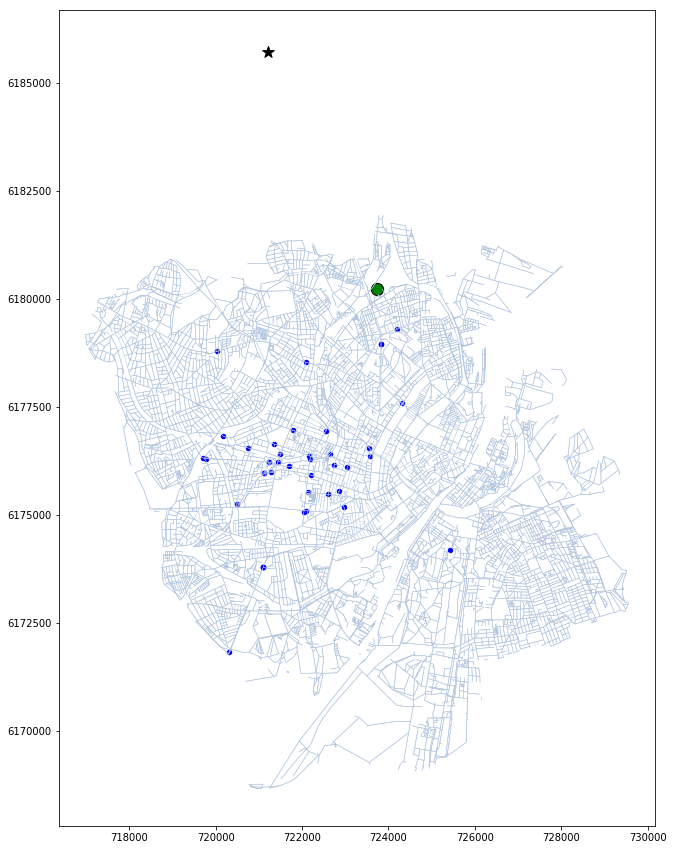}
    \caption{Map of the measurement stations and weather stations in Copenhagen}
    \label{fig:meas_st}
\end{figure}

Missing data are an issue for the wind speed data, where $37\%$ of the data were missing from the measurement station at DMI. The missing data are mainly centred around the winter months, potentially skewing the data representation and the model's ability to create accurate bicycle flow estimates. Therefore, we impute the missing data with wind speed observations from the nearest weather observation station, shown by the black asterisk in \Cref{fig:meas_st}. This station is approximately 7 kilometres removed from the DMI station. The imputed wind speeds only have $1.6\%$ missing data. The chosen approach may lead to a decreased precision in wind data for the bicycle volume estimation but presents a solid trade-off compared to missing $37\%$ of the data. 

\subsection{Bicycle flow estimation - model training and comparison}
To assess the performance of the proposed LSTMMDN to estimate historical hourly bicycle volumes, we compare it against other neural network architectures and, most importantly, the method based on calibration factors currently employed by the Danish government (the Seasonal Variation Factors, SVF). 

Four models for the estimation are being compared in various setups, the ANN setup proposed in \citet{Sekua2018EstimatingStudy} to estimate hourly car traffic in Maryland, LSTM and LSTMMDN in similar setups, and the SVF-based method used by the Danish government.

The ANN proposed by \citet{Sekua2018EstimatingStudy} contains three hidden layers of each 258 neurons with $20\%$ dropout in each layer and ELU activation \citep{Clevert2015FastELUs} in each hidden layer.
The LSTM networks have a single LSTM cell with $k=32$ or $k=64$ computation nodes in the LSTM-gates. The output from the LSTM cell is passed through a single layer with $m=6, m=8$ computational nodes, and finally to a single node output layer. We also evaluate the performance of an LSTM network where the LSTM cell is connected directly to the single output node.
The hidden layers all have linear $a(x) = x$ activation functions. 

The LSTMMDN is set up similarly to the described LSTM networks with $k=32$ or $k=64$ nodes in the LSTM-gates, and either $A=6$ or $A=8$ mixture components. The latter means that the output layer will have $m=A\times3$ as visualised in \Cref{fig:LSTMMDN}.

The described models are all "trained" using the training data set and the validation set used for monitoring and early stopping. After the model training, the test data set is used to compare model performance.  
The models are compared based on the following measures:
\begin{itemize}
\item The average mse ($\hat{mse}$) 
\item The average negative log-likelihood $-\log\hat{\mathcal{L}}$
\item The mse of the conditional average ($mse_{\mu}$)
\item The negative log-likelihood of the conditional average ($-\log \mathcal{L}_{\mu}$)
\end{itemize}
The first two are computed based on 100 posterior draws from the trained LSTMMDN, while the latter two are computed for the mean of the same 100 posterior draws. For the standard LSTM network, the ANN and the calibration factor method, the last two measures are computed for one forecast as they all estimate the conditional average cycling flow already.

The results are shown in \Cref{tab:comp_models}. 
Here $L(k)$ represents an LSTM cell with output dimension $k$, $\times \cdot m$ represents a connection with a hidden layer consisting of $m$ computation nodes, and  $ x G(A)$ refers to a connection with a GMM with $A$ mixture components and therefore an output layer with $A\times 3$ computational nodes. 

\begin{table}[htb]
    \centering
    \caption{Goodness-of-fit (GOF) measures related to various LSTMMDN structures and the SVFs, determined on the test data.}
    \begin{tabular}{l|c c c c c}
       model specification & $-\log\mathcal{L}_{\mu}$ & $-\log\hat{\mathcal{L}}$ & $MSE_{\mu}$ & $\hat{MSE}$ & Trainable parameters\\
       
       ANN: \citep{Sekua2018EstimatingStudy} & 5772 & - & 0.108 & - & 162,025\\
       LSTM: $L(k=32)$ & 5851 & - & 0.129 & - & 6,561\\
       LSTM: $L(k=64)$ & 5792 & - & 0.114 & - & 21,313\\
       LSTM: $L(k=32)x6$ & 5832 & - &  0.128 & - & 6,733 \\
       LSTM: $L(k=64)x6$ & 5777 & - & 0.110 & - & 21,645\\
       LSTMMDN: $L(k=32) x G(A=6)$ & 5805 & 5929 & 0.119 & 0.226 & 7,122\\
       LSTMMDN: $L(k=32) x G(A=8)$ & 5811 & 5933 & 0.121 & 0.234 & 7,320\\
       LSTMMDN: $L(k=64) x G(A=6)$ & 5753 & 5864 & 0.102 & 0.195 & 22,418\\
       LSTMMDN: $L(k=64) x G(A=8)$ & 5777 & 5872 & 0.109 & 0.201 & 22,808\\
      SVF-based estimates & 6570 & - &  0.377& -& -\\ 
    \end{tabular}
    \label{tab:comp_models}
\end{table}

Based on the GOF measures presented in \Cref{tab:comp_models}, the LSTMMDN  $L(k=64) x G(A=6)$ is the superior of the models. 
Noteworthy is that all the LSTMMDNs outperform their LSTM network counterpart with similar setups. 
Considering the less refined nature of the ANN \citep{Sekua2018EstimatingStudy} compared to the LSTM based models, it is surprising that it is the second-best performing model on the test data. However, the ANN has $\approx 8$ times as many estimable parameters as the second-largest model and takes significantly longer to train. 

The most relevant comparison is the model-based approaches vs the calibration-factor (SVF) method currently used in road agencies and the Danish Road Directorate.
Comparing all models-based estimates with the SVF-based estimates shows their superior performance when estimating the hourly bicycle flow while accounting for varying weather and time-related effects. With the improvements ranging from $66\% - 77\%$

For the remainder of this paper, we continue with the best performing LSTMMDN: $L(k=64) \times G(A=6)$, which will be compared further comparisons to the SVF-method for estimating hourly bicycle traffic. 

\subsection{LSTMMDN vs. SVF-calibration method}
\begin{figure}[htb]
    \centering
    \includegraphics[height=6cm, trim=0 0 0 0, clip]{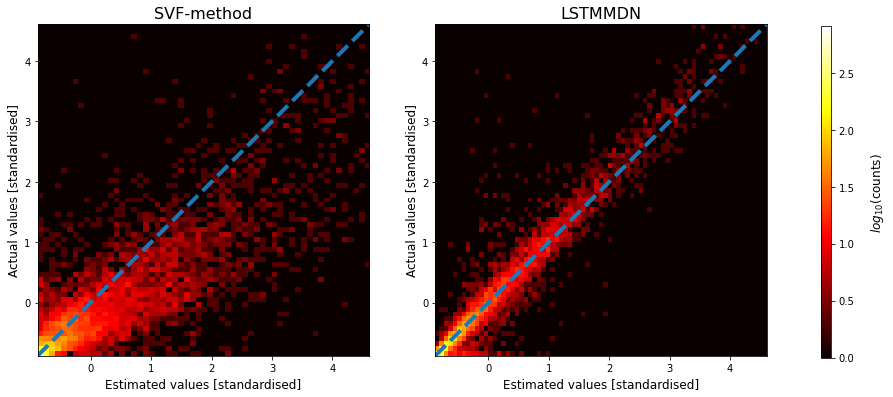}
    \caption{Heat maps comparing the prediction accuracy of the SVF method (left) and the LSTMMDN method (right) based on the reserved test data set.}
    \label{fig:heat}
\end{figure}

The LSTMMDN: $LSTM(64)\times G(6)$ performance on the test data yields an $MSE_{\mu}$ which is $\approx 77\%$ lower than that of the SVF method. To delve further into their respective ability to estimate hourly bicycle traffic accurately, we compare the fitted vs. actual bicycle flow using a heat plot in \Cref{fig:heat}.
The two heat plots reveal the superior ability of the LSTMMDN to estimate the hourly bicycle flow more accurately, as the concentration of estimates is much closer to the $45^{\circ}$ line establishing a $x=y$ relation. It is similarly apparent from the heatplots in \Cref{fig:heat} that the SVF method tends to overestimate the bicycle volumes compared to the actual bicycle volumes.

\begin{figure}[ht]
    \centering
   \includegraphics[width = 16cm]{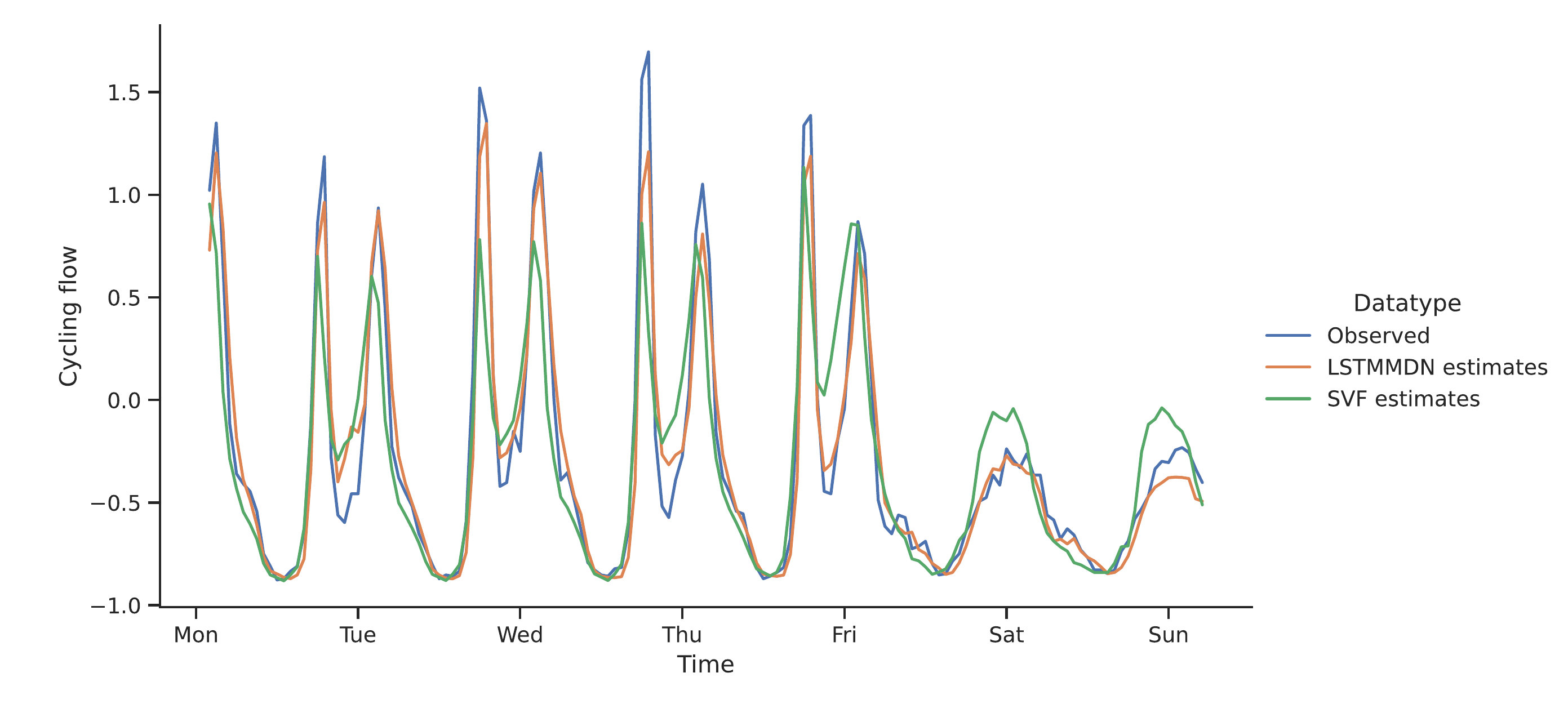}
    \caption{Visualisation of the standardised bicycle flow over 148 continuously registered hours from a randomly picked counting station. Blue: Actual counts of bicyclists from random counting station, orange: LSTMMDN cycling flow estimates on the same road section and week, and green: SVF-based cycling flow estimates for same road section and week.}
    \label{fig:TS}
\end{figure}

To look further into the discrepancies between the two bicycle flow estimation methods, we compare their representation of the bicycle flow over a continuous week. The plot in \Cref{fig:TS} shows a direct comparison of the standardised bicycle flow estimates from the LSTMMDN (orange) and the SVFs (green) for a continuous week at a monitoring station. The actual observations for the week at the station are shown in blue. \Cref{fig:TS} allows for a more fine-tuned diagnostic of the individual time-related effects learned by the LSTMMDN model vs the SVF calibration. Overall, the LSTMMDN estimates follow the observed values for the chosen road during this period much more tightly than the SVF estimates, coming closer in both the respective peaks and midday dips. However, the most significant difference between the LSTMMDN and the SVF estimates in \Cref{fig:TS} is the discrepancy of the SVF-method to estimate the traffic at weekends accurately. 

\subsection{Impact of bicycle exposure on bicycle accident analyses}
Having shown that our proposed method for estimating historical bicycle flow is superior to the calibration-factor method, we wish to assess the improved exposure data's impact on some city planners' and road agencies' tasks.
One such task is to improve cycling safety, which involves accident analysis. As previous studies have highlighted the necessity of accounting for exposure in accident analysis \citep{Vandenbulcke2014PredictingApproach, Thomas2013TheLiterature,Aldred2018CyclingLimits, Norros2016}, we wish to quantify the impact of the quality of the exposure data further. 

To do so, we estimate separate city-wide crash frequency models for Copenhagen with different exposure variables. Except for the exposure variable, all other exogenous variables are constant and equal across the different models. We consider three exposure variables: AAWCT, SVF-based hourly volume estimates, and LSTMMDN hourly volume estimates.
The crash frequency model is a simple Poisson regression. This model has been used in many previous studies involving crash frequency \citep{Lord2010TheAlternatives}. 

We consider for the response variable the aggregated amount of bicycle crashes in Copenhagen during any given hour in the period 2017-2020. As exogenous variables we consider the following:
Visibility $[m]$, Temperature $<0^{\circ}C$, Temperature $>20^{\circ}C$, Morning peak hours (7-9 weekday), Afternoon peak (15-17 weekday), Wind speed $<5m/s$, Wind speed $>9m/s$, Precipitation ($>0mm$), Bank holidays, Bicycle flow.

Hourly cycling exposure is estimated for the four years for each bicycle counting station using the LSTMMDN and calibration factors, respectively. Subsequently, the estimates are aggregated to approximate city-wide cycling exposure (similar to \citet{Dozza2017CrashData}). When using the AAWCT as an exposure variable, the volume is scaled to be the annual average hourly cycling traffic.

Due to missing weather data over the four years, leading to missing bicycle flow estimates, we undersample the bicycle flow to have three full years of data. The three years of data include 2104 bicycle crashes in 26,232 hours, meaning $0.08$ accidents per hour. 
Some summary statistics of the data used in the bicycle crash frequency model are shown in \Cref{tab:Poiss_data}.

\begin{table}[htb]
    \centering
    \begin{tabular}{c|c}
        Variable & Mean\\ \hline
        Visibility & 27,519 m \\
        Bank holiday & 0.034\\
        Exposure(LSTMMDN) & 5430 cyclists/hour\\
        Exposure(Exposure) & 6051 cyclists/hour\\
        Exposure(AAWCT) & 6079 cyclists/hour\\
        Morning Peak/ Afternoon peak & 0.089\\
        Temperature $< 0^{\circ}C$ &     0.032\\
        Temperature $> 20^{\circ}C$ &     0.075\\
        Wind speed $< 5 m/s$ &   0.102 \\
        Wind speed $> 9 m/s$&    0.008\\
        Precipitation $> 0 mm$  & 0.429\\
        \hline
    \end{tabular}
    \caption{Summary statistics of the data used in the Poisson model}
    \label{tab:Poiss_data}
\end{table}

The resulting parameter estimates for the estimated crash models are shown in \Cref{tab:Poisson}.

\begin{table}[htb]
    \centering
\resizebox{\textwidth}{!}{\begin{tabular}{lrr|rr|rr}
{} &\multicolumn{2}{c}{Model 1 (AAWCT based exposure)}   &  \multicolumn{2}{c}{Model 2 (SVF-based exposure)} &  \multicolumn{2}{c}{Model 3 (LSTMMDN based exposure)} \\
\hline
No. Obs        &                      $26,232$ & &                                             $26,232$ &&                                         $26,232$ \\
Estimated parameters & 10 && 10 && 10\\
log-likelihood &                      $-7,142$ &&                                              $-6,874$ & &                                        $-6,740$ \\
Deviance       &                      $10,287$ &  &                                             $9,750$ &    &                                      $9,483$ \\
$ \chi^{2}$     &                      $27,524$ &&                                              $27,704$ & &                                        $29,810$ \\
\hline \\
{} &  Parameter estimates &  p-value &  Parameter estimates &  p-value &  Parameter estimates &  p-value \\
Variables                                          &                     &           &                     &           &                     &           \\
\hline
Intercept &             $-23.390$ &     $<0.001$ &              $-7.958$ &     $<0.001$ &             $-10.740$ &     $<0.001$ \\
Visibility (for one log change)&               $0.071$ &     $0.029$ &              $-0.068$ &     $0.037$ &              $-0.062$ &     $0.055$ \\
Bank Holiday&              $-0.857$ &     $<0.001$ &              $-0.809$ &     $<0.001$ &              $-0.096$ &     $0.609$ \\
log(Exposure)&               $2.267$ &     $<0.001$ &               $0.697$ &     $<0.001$ &               $1.006$ &     $<0.001$ \\
Morning peak &               $1.098$ &     $<0.001$ &               $0.210$ &     $0.002$ &               $0.247$ &     $<0.001$ \\
Afternoon peak &               $1.194$ &     $0.003$ &               $0.296$&     $<0.001$ &               $0.284$ &     $ <0.001 $ \\
Temperature $< 0^{\circ}C$ &              $-0.512$ &     $<0.001$ &             $ -0.113$ &     $0.519$ &              $-0.027$ &     $0.877$ \\
Temperature $> 20^{\circ}C$ &               $0.422$ &     $0.774$ &               $0.075$ &     $0.270$ &               $0.224$ &     $40.001$ \\
Wind speed $< 5 m/s$&              $-0.023$ &     $0.946$ &               $0.181$ &     $0.025$ &              $0.136$ &     $0.090$ \\
Wind speed $> 9 m/s$ &              $-0.018$ &     $0.001$ &              $-0.058$ &     $0.824$ &              $-0.030$ &     $0.908$ \\
Precipitation &               $0.147$ &       $0.001$ &               $0.095$ &     $0.031$ &               $0.124$ &     $0.005$ \\
\hline
\end{tabular}
}
    \caption{Estimates for three Poisson regressions using similar input variables with the exception of varying the exposure variable (AAWCT, SVF-based hourly cycling volume and LSTMMDN-based hourly cycling volume).}
    \label{tab:Poisson}
\end{table}

We see from the results in \Cref{tab:Poisson} that Model 3 is the best based on the log-likelihood and Deviance. This leads us to conclude that the LSTMMDN estimates as exposure variables lead to superior model performance in crash risk analysis. The resulting log-likelihood is $5.5\%$ higher in Model 3 than in Model 1. 
Having changed only the exposure variable across the three models, this highlights that the quality of exposure estimates used in models have a substantial impact on fit to data. 

We also find that Model 2 (SVF based hourly cycling volumes) fits the data more accurately than Model 1, with a $3.9\%$ higher log-likelihood.

The results in \Cref{tab:Poisson} also reveal magnitude differences in parameter estimates and a difference in sign and significance of variable effects across the three models. 
We find that conclusions regarding various variable impacts on the bicycle frequency would vary depending on which of the three model variations is employed. For example, both low temperatures and high wind speeds would lower the crash risk in Model 1, contrasting the findings of the other two models. Further examples include lower wind speeds, which is significant in Model 2 at $5\%$.

\section{Discussion}
The current approach used by transport agencies to estimate hourly bicycle traffic relies on calibration factors and thus does not reflect variations in cycling exposure related to weather and other effects. This study aims to amend this issue by applying an LSTMMDN to estimate city-wide hourly bicycle volumes in Copenhagen based on the mean- daily traffic while accounting for weather and temporal effects. 
The results clearly show that the proposed LSTMMDN produces significantly more accurate estimates of hourly cycling flow than the Danish Road Directorate's calibration factor method. Conditional on the size of the applied network, the LSTMMDN yields $66\%$ to $77\%$ more accurate estimates of the hourly bicycle volume. 
The LSTMMDN contrasts models previously used in estimation and short-term traffic forecast and traffic estimation studies \citep{Ma2015LongData, Sekua2018EstimatingStudy} as it estimates a conditional cycling distribution, compared to only estimating a conditional average cycling flow. As such, the LSTMMDN should provide a more realistic representation of cycling. As cycling flows are treated as draws from the conditional cycling distribution, there is randomness in the system. This means that cycling flows on otherwise measurably identical roads will be different. This would not be the case for models estimating conditional averages.

Several focus areas of transport agencies could be assumed to be impacted by the improved bicycle volume estimates. With one such area being cycling safety improvement, we quantify the potential impact of improving the quality of bicycle flow estimation. Specifically, we estimate three bicycle crash frequency models, which are identical except for the exposure variable. The results show that improving the accuracy of bicycle volume estimates will result in better crash risk models. The best results are achieved from best to worst using the LSTMMDN estimates, the SVF estimates, and the modified mean daily aggregated cycling. These results add to previous research arguing that the accuracy of analyses is improved by including exposure, as opposed to no exposure \citep{Thomas2013TheLiterature, Norros2016}, by arguing that the disaggregation and quality of the exposure variable affect the accuracy of inference. A potentially worrying notion was raised from the results of the crash frequency models. We observe both sign and significance level changes concerning the variable effects when comparing the models in \Cref{tab:Poisson}. This is evidence of aggregation bias influencing the estimated variable effects when using aggregated exposure estimates and raises questions about the validity of results from studies using highly aggregated exposure in models. However, a definitive clarification would require an in-depth analysis and is the subject of future studies. Nonetheless, it presents a strong argument for the need for further bicycle observation/monitoring efforts and the methods for estimating bicycle flow.

\subsection{Limitations}
\subsubsection{Model-based hourly cycling volumes}
The model presented in the current study estimates bicycle flow based on limited information. It offers an alternative to the calibration method currently employed in transport agencies that relies on already present estimates of the mean daily traffic to estimate hourly bicycle traffic but also accounts for weather and time-dependent effect. This should provide a valuable tool to be linked with recently developed large-scale models that estimate link-based cycling volumes, but often describing the mean day \citep{Kjems2019COMPASS:Hovedstadsomradet, AledDavies2017Cynemon-CyclingPlanning}. However, several variables are not accounted for that could be considered when attempting to accurately estimate cycling flow.
Accounting for the built environment, network structure and connected routes could improve the model's accuracy. Applying models similar to \citet{Bao2019AData}, which combine network temporal (waather) features, could considered for future research.
Previous research suggests that cyclists prefer bicycle paths separated from motorised traffic \citep{Aldred2017CyclingAge, Broach2012WhereData, TransportforLondon2014AttitudesTfL}, and future applications should seek to include car traffic as a predictor in the model.
The specific strength of the LSTM is its ability to handle very long time-series data. In this study, however, the input sequences passed to the LSTM only contain sequences of six 10-minute weather intervals. Future research could concern itself with the topic and include extended series of weather data into the bicycle flow model and include previous bicycle counts to improve accuracy.

\subsubsection{Data}
The current model relies on the availability of monitoring of hourly bicycle volumes. Therefore, if no data on bicycle volumes is available other methods need to be explored. 
The increased availability of disrupting technologies to monitor cyclists their and behaviour via systems such as instrumented cyclists \citep{Gustafsson2013AArea, Roos2020IdentifieringCykelhjalmsdata} also implies the potential for new ways and methods to estimate cycling that potentially could turn cycling counting stations obsolete in the future. Meanwhile, the newer technologies enabling the monitoring of cyclist volumes and behaviour come from private companies and their products, making them potentially less viable for the transport agencies. Therefore, although sparse, the most easily accessible cycling monitoring data for transport agencies and safety engineers still stems from automated bicycle counting stations. Hence it should still be in their interest to develop models to improve the hourly bicycle estimates based on what little information about the cycling exposure is currently available.

\subsection{Bicycle Crash models}
The crash model and the variables used for those models are very simplistic and quantify the impact that better bicycle volume estimates can have on the accident models that are crucial to making informed decisions to increase cycling safety. Many studies investigate the factors related to bicycle crashes and the outcomes thereof \citep{Janstrup2019APrioritization, Myhrmann2021FactorsCrashes, Fountas2021AddressingMeans, Schepers2020TheMost,Aldred2018CyclingLimits, Kaplan2013Cyclist-motoristApproach.,Kim2007BicyclistAccidents}, and this study does not try to conduct a deep risk analysis of bicycle crashes.  
Nevertheless, the three Poisson regressions results indicate that the bicycle flow estimates affect the output. This result makes it plausible that the same would be the case in more advanced models. 

\section{Conclusion}
This paper focuses on estimating historical bicycle flows that transport agencies use for various tasks such as improving cycling safety. The proposed LSTMMDN estimates cycling traffic based on the mean daily traffic and accounts for weather and time-related factors. This method significantly outperforms the current calibration factor method to obtain hourly bicycle flow estimates, and the cycling flow estimates are up to $77\%$ more accurate in the studied example. The LSTMMDN also provides a more realistic cycling representation due to the model's built-in uncertainty.
Overall this suggests that traffic estimation efforts in transport agencies would benefit from accounting for weather and other influencing factors as well as embracing newer statistical frameworks. Also, since the quality of exposure data could easily be thought to impact the results of their analyses. 
In line with this, the current study quantifies the impact of the improved cycling flow data from the LSTMMDN in accident analysis. The results clearly show how these models, generally used to make informed decisions regarding mitigating action, benefit from the more disaggregated and accurate bicycle flow data. Lastly, the results also suggest that highly aggregated exposure data could lead to erroneous conclusions.

\section*{Declaration of Competing Interests}
The authors declare that they have no known competing financial interests or personal relationships that could have appeared to influence the work reported in this paper.

\section*{Acknowledgements}
We thank Filipe Rodrigues and Mads Paulsen for valuable feedback on an earlier version of the paper process.
\bibliography{references.bib}

\end{document}